\documentclass[11pt]{article}
\usepackage[margin=1in]{geometry}
\usepackage{booktabs}
\usepackage{amsmath}
\usepackage[hidelinks]{hyperref}
\usepackage{xcolor}
\usepackage{graphicx}
\usepackage{fvextra}

\title{What Moves a LoCoMo Score? Memory Context, Answer Prompts and Judges}
\author{Nikhil Kumar Jha \quad Amitoj Cheema\\ becosAI \\ \texttt{nikhil@becos.ai}, \texttt{amitoj@becos.ai}}
\date{November 2026}

\begin{document}
\maketitle

\begin{abstract}
LoCoMo is a widely reported benchmark for long-term conversational memory, but published
scores mix memory quality with choices of answer prompt, judge and baseline layout that are rarely
reported together. We measure how much these choices move a score, and evaluate Becos, a team memory
system for coding agents, as a black box under all of them: ten conversations (1,540 questions), two
answer models (gpt-4o-mini, gpt-4.1-mini), two answer prompts and three LLM judges, against a
whole-conversation baseline rerun under identical conditions. Changing only the judge moves a score by
up to 7.2 points and changing only the answer prompt by up to 9.3. A blind human check of 150 answers finds the
judges reliable when they agree, and the primary judge's errors on disputed answers working against
Becos rather than for it. Becos answers from a context with
90\% fewer characters than the whole conversation. Its configuration was developed on LoCoMo itself,
so there is no held-out set. With gpt-4o-mini answers it scores clearly higher
than the whole conversation (79.7 vs 73.0 with the short prompt, paired 95\% interval for the
difference +3.8 to +9.5; 88.8 vs 77.5 with a step-by-step prompt). With gpt-4.1-mini answers the two
score about the same (83.2 vs 83.1; 92.5 vs 92.1). We release every answer and verdict and the scoring
script.
\end{abstract}

\section{Introduction}
Memory systems for LLM agents are compared on a small number of public benchmarks, LoCoMo
\cite{locomo} chief among them. Reported scores have climbed from the low 60s \cite{mem0} to above 90
\cite{msft}, but papers differ in answer models, answer prompts, judges and even in how they lay out
the whole-conversation baseline, and rarely report all of these. That makes it hard to tell a better
memory from a different setup.

This paper is primarily a controlled study of that sensitivity, with a black-box evaluation of one
memory system alongside it. We rerun the whole-conversation baseline in-house, vary one setting at a
time, and report every combination.

\begin{figure}[t]\centering
\includegraphics[width=\linewidth]{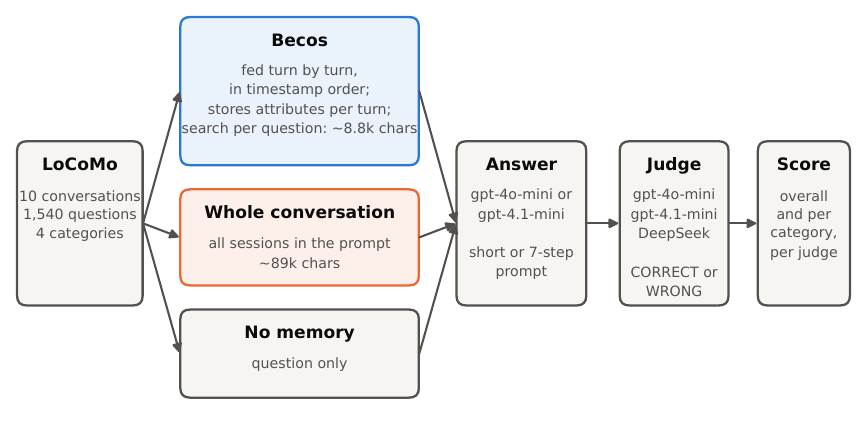}
\caption{Experiment overview. Each question is answered from one of three contexts, by one of two
answer models with one of two answer prompts, and every answer is graded by three judges.}
\label{fig:overview}
\end{figure}

Our contributions:
\begin{itemize}
  \item A sensitivity analysis on LoCoMo: answer model, answer prompt, judge, a date-generous judge,
        and the layout of the whole-conversation baseline (Section~\ref{sec:sensitivity}).
  \item A black-box evaluation of Becos on all ten conversations under 2 answer models $\times$ 2
        answer prompts $\times$ 3 judges, with paired uncertainty estimates
        (Figures~\ref{fig:overview} and \ref{fig:results}; Tables~\ref{tab:main}--\ref{tab:paired}).
  \item A consistency check of published LoCoMo tables (Section~\ref{sec:published}).
  \item The prompts, every answer and verdict, and the scoring script.
\end{itemize}

\section{Related work}
\paragraph{Memory for LLM agents.} MemGPT \cite{memgpt} manages a limited context window by moving
information between memory tiers, in the manner of an operating system. Later systems store
information extracted from conversations and retrieve it for each query: Mem0 \cite{mem0}, Zep's
time-aware knowledge graph \cite{zep}, A-Mem's linked, self-updating notes \cite{amem}, Nemori's memory
distillation \cite{nemori}, EverMemOS \cite{evermemos} and the curation and consolidation method of
\cite{msft}. Most of them report LoCoMo. We evaluate one such system as a black box and focus on the
evaluation choices around it.

\paragraph{Benchmarks.} Multi-Session Chat \cite{msc} introduced open-domain conversations that span
several sessions. LoCoMo \cite{locomo} has far longer conversations and evaluates question answering,
event summarization and multimodal dialogue; we use its question answering. LongMemEval \cite{longmemeval}
tests five long-term memory abilities of chat assistants over long interaction histories. An
independent audit of LoCoMo \cite{audit} documents errors in its answer key and the leniency of its
standard judge.

\paragraph{Long contexts.} The whole-conversation baseline depends on how well the answer model uses a
long input. Models use information in the middle of a long input less reliably than at its start or
end \cite{lostmiddle}, and many perform worse well before their stated context length \cite{ruler}.
Either effect could explain why gpt-4o-mini does better from a short context in our results; we do not
test which.

\paragraph{LLM judges.} LoCoMo scores are usually assigned by an LLM judge. Strong LLM judges agree with
human preferences about as well as humans agree with each other, but show position, verbosity and
self-enhancement biases \cite{judge}. Verbosity bias matters for our 7-step prompt, which produces
longer final answers (Section~\ref{sec:sensitivity}).

\section{Setup}

\subsection{Benchmark}
LoCoMo \cite{locomo} has ten long two-person conversations (369 to 689 messages over 19 to 32
sessions) with question--answer pairs in five categories. As in prior work \cite{mem0,nemori,msft} we
drop the adversarial category (446 questions about events the conversation attributes to someone else,
for which the data gives only the trap answer and no gold answer, so the standard judge cannot score
them), leaving 1,540 questions: 841 single-hop, 282 multi-hop, 321 temporal and 96 open-domain
(category ids 4, 1, 2 and 3 in the released data).

\subsection{Systems compared}
\paragraph{Becos.} Becos is a commercial memory system for teams using AI coding agents, developed by
the authors' company (\url{https://becos.ai}); we evaluate it as a black box. Only conversation histories are
ingested into Becos, one turn at a time, oldest session first, in timestamp order, the way an agent
session reaches it. Evaluation questions are supplied at retrieval time; gold answers are used only for
judging and scoring. At every turn Becos breaks the message down into
attributes (who, what, when) and stores them. To answer a question it searches that store and returns
a context of at most 8,950 characters, which can include short excerpts of the original messages. The
configuration evaluated here is a research configuration for conversation projects, frozen on 9
November 2026, and every Becos run reported here was made with it on that day (the baselines were run
from 7 November); it is not yet part of Becos for coding agents.

\paragraph{Whole conversation.} The entire conversation the question belongs to, in the answer prompt,
one line per message, with each session's date as a header line (the layout of the released data).
This is the standard baseline \cite{mem0,nemori,msft}; we rerun it under the same answer models,
prompts and judges.

\paragraph{No memory.} The question alone, as a floor.

\subsection{Development and evaluation}\label{sec:dev}
This work was a research effort, and Becos's configuration for conversation projects was developed on
LoCoMo itself. Changes to how Becos stores a conversation were made while examining errors on conv-26,
conv-30 and conv-41; retrieval settings were chosen on conv-42, conv-43 and conv-44. The configuration
was then frozen and run unchanged on all ten conversations. conv-47 to conv-50 (655 questions) were
never examined question by question, but their overall results under an earlier configuration were
seen before the last change was adopted, so there is no fully held-out set. We report results on all
ten and, separately, on conv-47 to conv-50 as the least-used conversations (Table~\ref{tab:paired});
all results should be read as an evaluation on a benchmark that was also used during development.

\subsection{Answering and judging}
\paragraph{Answer prompts.} \emph{Short}: Nemori's answer prompt \cite{nemori,nemoricode}, which asks
for an answer under 5--6 words. \emph{7-step}: the same prompt with that instruction replaced by seven
written reasoning steps (list relevant memories with timestamps, extract key details, connect facts,
convert relative times, resolve conflicts, check details, plan the answer) and a \texttt{FINAL ANSWER}
line with no length limit; the step list follows the LoCoMo audit's description \cite{audit} of
EverMemOS's chain-of-thought prompt \cite{evermemos}, in our own words. Only the text after
\texttt{FINAL ANSWER} is passed to the judge. Both are in Appendix~\ref{app:prompts}. Answer models are
gpt-4o-mini and gpt-4.1-mini at temperature 0, one pass per question; the API responses report the
snapshots gpt-4o-mini-2024-07-18 and gpt-4.1-mini-2025-04-14 for every call.

\paragraph{Judges.} Nemori's judge prompt \cite{nemoricode} with three judge models: gpt-4o-mini (the
judge Nemori states, and our primary column), gpt-4.1-mini, and DeepSeek-v4-flash (which the DeepSeek API reports as deepseek-flash). Each verdict is
CORRECT or WRONG; a score is the share of questions judged correct, so the overall score weights each
category by its number of questions. Judges are never averaged. All runs took place on 7--9 November
2026.

\section{Results}
\begin{figure}[t]\centering
\includegraphics[width=\linewidth]{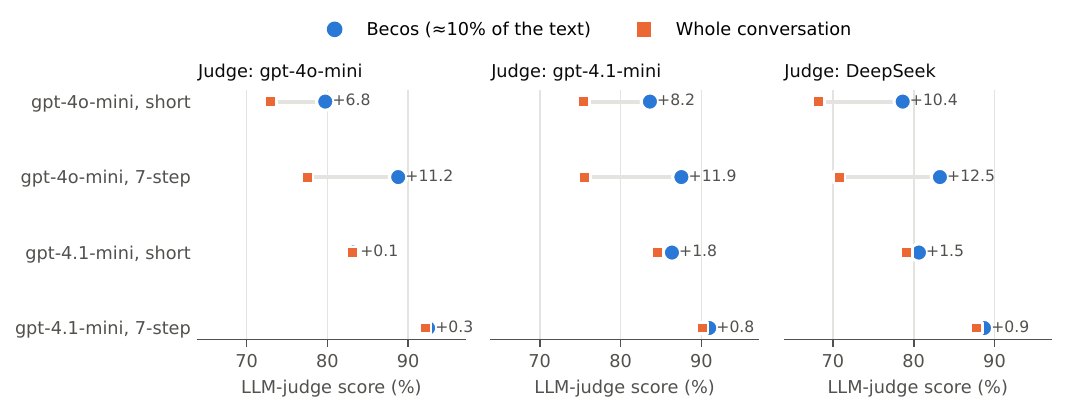}
\caption{Becos (circles) against the whole conversation (squares) on all 1,540 questions, for each
answer model and prompt (rows) and each judge (panels). Numbers give Becos's difference in points,
computed before rounding.}
\label{fig:results}
\end{figure}

\begin{table}[t]\centering\small
\begin{tabular}{lllrrr}
\toprule
Answer model & Answer prompt & Context & \multicolumn{3}{c}{Judge} \\
\cmidrule(l){4-6} & & & 4o-mini & 4.1-mini & DeepSeek \\
\midrule
gpt-4o-mini & short & No memory & 19.5 & 24.5 & -- \\
gpt-4o-mini & short & Whole conversation & 73.0 & 75.5 & 68.2 \\
gpt-4o-mini & short & Becos & \textbf{79.7} & \textbf{83.6} & \textbf{78.6} \\
\midrule gpt-4o-mini & 7-step & Whole conversation & 77.5 & 75.6 & 70.8 \\
gpt-4o-mini & 7-step & Becos & \textbf{88.8} & \textbf{87.5} & \textbf{83.2} \\
\midrule gpt-4.1-mini & short & Whole conversation & 83.1 & 84.6 & 79.2 \\
gpt-4.1-mini & short & Becos & \textbf{83.2} & \textbf{86.4} & \textbf{80.6} \\
\midrule gpt-4.1-mini & 7-step & Whole conversation & 92.1 & 90.1 & 87.8 \\
gpt-4.1-mini & 7-step & Becos & \textbf{92.5} & \textbf{91.0} & \textbf{88.7} \\
\bottomrule
\end{tabular}
\caption{All ten conversations (1,540 questions). LLM-judge score (\%) by answer model, answer prompt,
context and judge. The no-memory row was not judged by DeepSeek.}
\label{tab:main}
\end{table}

\begin{table}[t]\centering\small
\begin{tabular}{llrrrrr}
\toprule
& & \multicolumn{3}{c}{All ten conversations (1,540)} & \multicolumn{2}{c}{conv-47 to conv-50 (655)} \\
\cmidrule(l){3-5}\cmidrule(l){6-7}
Answers & Prompt & Extra correct & Difference & 95\% interval & Becos & Whole \\
\midrule
gpt-4o-mini & short & +104 & +6.8 & [+3.8, +9.5] & 78.8 & 70.5 \\
gpt-4o-mini & 7-step & +173 & +11.2 & [+9.2, +13.1] & 89.3 & 77.6 \\
gpt-4.1-mini & short & +1 & +0.1 & [-2.3, +2.2] & 83.2 & 82.3 \\
gpt-4.1-mini & 7-step & +5 & +0.3 & [-0.6, +1.1] & 93.4 & 93.3 \\
\bottomrule
\end{tabular}
\caption{Becos minus the whole conversation, gpt-4o-mini judge. ``Extra correct'' counts the
additional questions judged correct. The 95\% interval is a paired bootstrap that resamples whole
conversations (20,000 replicates), conditional on the saved answers and verdicts; with ten
conversations it is approximate. The last two columns are the four conversations least used during
development (Section~\ref{sec:dev}).}
\label{tab:paired}
\end{table}

\paragraph{Clear gains with gpt-4o-mini; similar scores with gpt-4.1-mini at a tenth of the context.}
Figure~\ref{fig:results}, Tables~\ref{tab:main} and \ref{tab:paired}. With gpt-4o-mini answers Becos
scores higher than the whole conversation under every judge: by 6.8, 8.2 and 10.4 points with the short
prompt and by 11.2 to 12.5 with the 7-step prompt. The paired interval under the gpt-4o-mini judge
excludes zero (+3.8 to +9.5 short, +9.2 to +13.1 7-step), and the four conversations least used in
development show a similar gap (78.8 vs 70.5; 89.3 vs 77.6). With gpt-4.1-mini answers the two score
about the same: Becos answers 1 more question correctly than the whole conversation with the short
prompt and 5 more with the 7-step prompt, out of 1,540, and both intervals include zero. Becos reaches
these scores from about a tenth of the context (Section~\ref{sec:size}).

\paragraph{By category.} Table~\ref{tab:cat}. Becos's largest margin is on temporal questions (79.8 vs
50.5 with gpt-4o-mini answers, short prompt). Becos stores each event with its date, while a model
reading the whole conversation must find the right session header and do the date arithmetic; part of
this gap depends on that baseline layout (Section~\ref{sec:sensitivity}). Becos's weakest category is
open-domain, which needs world knowledge on top of the conversation. With the short prompt the whole
conversation is slightly ahead on single-hop; with the 7-step prompt the two are level (95.4 vs 95.8
with gpt-4.1-mini answers).

\begin{table}[t]\centering\footnotesize\setlength{\tabcolsep}{4pt}
\begin{tabular}{lllrrrrr}
\toprule
Answers & Prompt & Context & Single-hop & Multi-hop & Temporal & Open-domain & Overall \\
\midrule
4o-mini & short & Whole conv. & 85.9 & 65.6 & 50.5 & 57.3 & 73.0 \\
4o-mini & short & Becos & 85.4 & 72.3 & 79.8 & 52.1 & 79.7 \\
\midrule 4o-mini & 7-step & Whole conv. & 86.7 & 78.7 & 56.1 & 65.6 & 77.5 \\
4o-mini & 7-step & Becos & 92.7 & 86.5 & 86.0 & 69.8 & 88.8 \\
\midrule 4.1-mini & short & Whole conv. & 88.9 & 80.1 & 78.5 & 56.2 & 83.1 \\
4.1-mini & short & Becos & 87.8 & 83.0 & 76.9 & 64.6 & 83.2 \\
\midrule 4.1-mini & 7-step & Whole conv. & 95.8 & 92.2 & 87.2 & 76.0 & 92.1 \\
4.1-mini & 7-step & Becos & 95.4 & 91.1 & 90.0 & 79.2 & 92.5 \\
\bottomrule
\end{tabular}
\caption{By category, gpt-4o-mini judge. Questions: 841 single-hop, 282 multi-hop, 321 temporal, 96
open-domain.}
\label{tab:cat}
\end{table}

\paragraph{Per conversation.} Table~\ref{tab:conv}. With gpt-4o-mini answers and the short prompt
Becos is ahead on eight of the ten conversations; with gpt-4.1-mini answers and the 7-step prompt it is
level or ahead on seven and within 4 points on the other three. Per-conversation differences rest on
81 to 199 questions each and vary in both directions.

\begin{table}[t]\centering\small
\begin{tabular}{lrrrrrrr}
\toprule
& & \multicolumn{2}{c}{4o-mini, short} & \multicolumn{2}{c}{4.1-mini, short} & \multicolumn{2}{c}{4.1-mini, 7-step} \\
\cmidrule(l){3-4}\cmidrule(l){5-6}\cmidrule(l){7-8}
Conversation & Qs & Whole & Becos & Whole & Becos & Whole & Becos \\
\midrule
conv-26 & 152 & 74.3 & 85.5 & 86.8 & 90.1 & 97.4 & 99.3 \\
conv-30 & 81 & 69.1 & 76.5 & 86.4 & 75.3 & 95.1 & 91.4 \\
conv-41 & 152 & 76.3 & 86.8 & 88.8 & 88.2 & 94.1 & 92.1 \\
conv-42 & 199 & 69.8 & 77.4 & 77.4 & 78.9 & 87.4 & 88.4 \\
conv-43 & 178 & 78.1 & 77.0 & 82.6 & 81.5 & 86.5 & 88.8 \\
conv-44 & 123 & 80.5 & 78.9 & 83.7 & 82.9 & 91.1 & 91.9 \\
conv-47 & 150 & 67.3 & 80.7 & 78.0 & 84.7 & 92.7 & 93.3 \\
conv-48 & 191 & 75.4 & 80.1 & 83.2 & 84.3 & 93.2 & 93.7 \\
conv-49 & 156 & 67.3 & 76.3 & 84.6 & 79.5 & 92.9 & 92.3 \\
conv-50 & 158 & 70.9 & 77.8 & 82.9 & 84.2 & 94.3 & 94.3 \\
\bottomrule
\end{tabular}
\caption{Per conversation, gpt-4o-mini judge. ``Whole'' is the whole conversation.}
\label{tab:conv}
\end{table}

\section{Context size}\label{sec:size}
Every question gets the whole conversation it belongs to, so the whole-conversation context per
question averages 89,048 characters over the 1,540 questions (51,927 to 105,427 per conversation).
Becos's context averages 8,774 characters (6,226 to 8,950), 9.9\% as much (90.1\% fewer characters),
and does not grow with the length of the conversation. In tokens, as counted by the API for the short
prompt: the context is 2,337 tokens with Becos against 20,525 with the whole conversation (88.6\%
fewer), and the whole answer prompt, including 480 tokens of instructions and question, is 2,817
against 21,005 (86.6\% fewer). These counts cover answering only; Becos's own cost of storing and
searching the conversation is not included.

\section{What else moves the score}\label{sec:sensitivity}
\paragraph{The judge.} On identical answers the three judges differ by up to 7.2 points: the whole
conversation with gpt-4o-mini answers scores 75.5 under the gpt-4.1-mini judge and 68.2 under
DeepSeek (73.0 under gpt-4o-mini; Table~\ref{tab:main}). The judges agree on 90--93\% of individual
verdicts for short answers and 95--97\% for 7-step answers. DeepSeek is the strictest, and stricter on
the whole conversation than on Becos.

\paragraph{A human check of the judges.} One author graded 150 answers (gpt-4o-mini answers, short
prompt) blind to system and judge, as correct, wrong or cannot tell, against a written rubric: 60
answers from each system on which the three judges do not all agree (of 176 for Becos and 200 for the
whole conversation) and 15 from each on which they all agree. The 17 marked cannot tell (13 of them
temporal) are excluded. Where all three judges agree, each matches the human verdict on 29 of 30
answers. Where they disagree, gpt-4.1-mini matches it on 65 of 103, DeepSeek on 57 and gpt-4o-mini,
our primary judge, on 47. The primary judge's errors do not favour Becos: on Becos's answers it was
too strict 22 times and too lenient 7, on the whole conversation's too strict 12 times and too
lenient 15. Keeping the unanimous verdicts and applying the human rate of correct answers to each
system's disputed questions gives about 82.2 for Becos and 71.4 for the whole conversation, a
difference of 10.7 points before rounding (8.4 to 13.0 when resampling only the graded answers),
against the judge's 6.8.
With one grader, who is also an author, we read this as a check that judge error does not explain
Becos's lead, not as a corrected score.

\paragraph{The answer prompt.} Switching from the short prompt to the 7-step prompt, with nothing else
changed, raises scores by 4.6 to 9.3 points under the gpt-4o-mini judge and by 0.1 to 8.6 under the
other two (Table~\ref{tab:main}). The 7-step prompt changes three things at once: it adds written
reasoning steps, removes the short-answer limit (median final answers grow from 3--5 words to 17--29),
and has only the final answer judged. Our comparison does not separate these. One hypothesis is that
part of the gain is genuine---longer answers keep the specific detail a question asks for---and part
is grading: the judge prompt asks the judge to be generous, and an independent audit found that this
judge accepts 63\% of deliberately vague but on-topic wrong answers \cite{audit}. A length-controlled
ablation or a human audit would be needed to settle it. We therefore report both prompts side by side
and do not compare a 7-step number with a short-answer number.

\paragraph{Date leniency.} A diagnostic judge that also accepts dates within seven days of the gold
date raises the whole conversation's temporal score from 50.5 to 58.3 and Becos's from 79.8 to 81.9
(gpt-4o-mini answers). We do not report it as a score.

\paragraph{Layout of the whole-conversation baseline.} Putting the session date on every message
instead of once per session raises the whole conversation on conv-47 from 67.3 to 73.3 (temporal 26.5
to 44.1). We ran this on one conversation only, so its effect across the benchmark is unknown. Our
comparisons are therefore against the baseline as laid out in the released data, not against every
possible layout; baselines in different papers may differ in this way without saying so.

\section{Comparison with published results}\label{sec:published}
\begin{table}[t]\centering\small
\begin{tabular}{llrl}
\toprule
Answers & System & LLM score & Judge \\
\midrule
4o-mini & Mem0 \cite{nemori} & 61.3 & 4o-mini \\
4o-mini & Zep \cite{nemori} & 58.5 & 4o-mini \\
4o-mini & LangMem \cite{nemori} & 51.3 & 4o-mini \\
4o-mini & Full context \cite{nemori} & 72.3 & 4o-mini \\
4o-mini & Nemori \cite{nemori} & 73.0 & 4o-mini \\
4o-mini & Whole conversation (ours, short) & 73.0 & 4o-mini \\
4o-mini & \textbf{Becos (ours, short)} & \textbf{79.7} & 4o-mini \\
4o-mini & Hybrid curation \cite{msft} & 83.5 & not stated \\
4o-mini & Whole conversation (ours, 7-step) & 77.5 & 4o-mini \\
4o-mini & \textbf{Becos (ours, 7-step)} & \textbf{88.8} & 4o-mini \\
\midrule
4.1-mini & Mem0 \cite{nemori} & 66.3 & 4o-mini \\
4.1-mini & Zep \cite{nemori} & 61.6 & 4o-mini \\
4.1-mini & LangMem \cite{nemori} & 73.4 & 4o-mini \\
4.1-mini & Full context \cite{nemori} & 80.6 & 4o-mini \\
4.1-mini & Nemori \cite{nemori} & 80.8 & 4o-mini \\
4.1-mini & Whole conversation (ours, short) & 83.1 & 4o-mini \\
4.1-mini & \textbf{Becos (ours, short)} & \textbf{83.2} & 4o-mini \\
4.1-mini & Hybrid curation \cite{msft} & 90.2 & not stated \\
4.1-mini & Whole conversation (ours, 7-step) & 92.1 & 4o-mini \\
4.1-mini & \textbf{Becos (ours, 7-step)} & \textbf{92.5} & 4o-mini \\
\bottomrule
\end{tabular}
\caption{Overall LLM-judge score next to published results on all ten conversations. Published
numbers are from Nemori's Table~2 \cite{nemori} (judge gpt-4o-mini) and from \cite{msft} (judge and
answer prompt not stated). Our rows use the gpt-4o-mini judge. Matching the judge model does not make
these controlled comparisons: answer prompts and other settings may differ.}
\label{tab:published}
\end{table}

Table~\ref{tab:published} sets our results next to published ones. Nemori states its judge
(gpt-4o-mini), so its table is the closest to ours: Becos with the short prompt scores 79.7 against
73.0 reported for Nemori with gpt-4o-mini answers, and 83.2 against 80.8 with gpt-4.1-mini, and above
the Mem0, Zep and LangMem results in the same table. Our whole-conversation baseline lands within 1--3
points of Nemori's (73.0 vs 72.3; 83.1 vs 80.6).

The hybrid curation method of \cite{msft} reports 83.5 and 90.2, with a judge and answer prompt the
paper does not state. Its whole-conversation baseline (79.0, 91.3) is higher than short-answer
baselines (ours 73.0 / 83.1; Nemori 72.3 / 80.6) and similar to what the LoCoMo audit obtains with a
chain-of-thought prompt (79.9, 92.6) \cite{audit}, but similar scores do not establish matching
settings. We list its results for context; their comparability with ours is unknown pending the
authors' clarification.

\paragraph{A consistency check.} An overall score over the 1,540 questions equals the
question-weighted mean of the four category scores; our tables and Nemori's satisfy this exactly
(73.0 = 73.0; 72.3 = 72.3). The per-category whole-conversation figures printed in \cite{msft} give
76.2 and 89.5 rather than the reported 79.0 and 91.3. Of the 24 ways to assign the four printed columns
to the four categories, exactly one brings both rows within 0.15 points of the reported overall (78.89
vs 79.0; 91.29 vs 91.3), substantially reducing the discrepancy, though the first row still falls just
outside what rounding alone explains. Under that assignment the column printed as temporal corresponds
to single-hop, single-hop to multi-hop, and multi-hop to temporal, and the baseline's temporal score
would be 53.2 and 87.8, in line with ours and Nemori's, instead of 90.3 and 94.8. This is the pattern
produced by mapping LoCoMo's numeric category ids in a different order, but we have not confirmed the
cause. We contacted the authors on 9 November 2026 and had not received a reply at the time of
writing.

\section{Limitations}
\begin{itemize}
  \item The configuration was developed on LoCoMo (six of the ten conversations informed specific
        changes; Section~\ref{sec:dev}), and there is no fully held-out set. The four least-used
        conversations show the same pattern, but LoCoMo is not an unseen benchmark for this
        configuration.
  \item One answer pass per question, and ten conversations; the paired intervals are approximate and
        conditional on the saved answers and verdicts.
  \item The whole-conversation baseline uses the released data's layout; a date on every message
        helped it on the one conversation we tried (Section~\ref{sec:sensitivity}).
  \item LoCoMo is personal chat, not the coding work Becos is built for. These results do not show
        how Becos does on coding work.
  \item LoCoMo's answer key has errors \cite{audit}, and its judge is lenient; both cap what a score
        can mean.
  \item The 7-step prompt changes several things at once; its gains are not attributed to reasoning
        alone.
  \item Each conversation was loaded into Becos once; loading again may store slightly different
        attributes. Retrieval was checked to return identical text across two runs.
  \item Becos uses an LLM when storing each turn, and that model is not one of the answer models.
        Part of Becos's gain with gpt-4o-mini answers may come from a different model having
        processed the conversation at storage time.
  \item The human check has one grader, an author, on 150 answers from one setting; a second
        grader and a larger sample would be needed to report human-judged scores.
  \item Context sizes and token counts cover answering only; Becos's own cost and latency of storing
        and searching are not measured.
\end{itemize}

\section*{Competing interests}
Both authors work at becosAI, which develops Becos. All answers and verdicts, and the scripts to rerun
the baselines, are released so the comparison can be checked independently of us.

\section{Reproducibility}
We release at \url{https://github.com/becos-ai/locomo-results} every question's answer
and every judge verdict behind the tables, the human-check sample with its key and grades, and \texttt{locomo\_eval.py}, a standalone script (Python
standard library only) that answers LoCoMo questions with the whole conversation, no context, or any
system's own retrieved context, under both prompts of Appendix~\ref{app:prompts}, judges answers, and
scores them overall and by category. The release supports three levels of checking: \emph{score
reproduction}---rescoring our released verdicts reproduces every number in
Tables~\ref{tab:main}--\ref{tab:paired}, including the paired intervals (a \texttt{paired} command
with the same seed), and our answers can be re-judged with any judge;
\emph{baseline rerunning}---the whole-conversation and no-memory runs can be regenerated end to end;
\emph{memory-system reproduction}---not possible from the release, since Becos's retrieved contexts
and implementation are not included. Answers and verdicts are released under CC BY-NC 4.0 (as LoCoMo
is), the script under MIT. LoCoMo's conversations were generated by LLM agents, so no personal data is
involved. Researchers can request API access to Becos from the authors. Total API cost
of the runs reported here: about \$63 of OpenAI calls (answers and the two OpenAI judges) and \$10.31 of
DeepSeek calls (the third judge and loading).

\appendix
\section{Prompts}\label{app:prompts}
We use four prompts. Three are Nemori's evaluation prompts, reproduced verbatim from the
nemori-ai/nemori repository at commit d2a6dff, directory \texttt{evaluation/locomo} (MIT licence)
\cite{nemoricode}: the short answer prompt from \texttt{search.py}, and the judge's system message and
judge prompt from \texttt{metrics/llm\_judge.py}. The fourth, the 7-step answer prompt, is ours. It is
Nemori's answer prompt with instruction 8 (``less than 5-6 words'') replaced by seven reasoning steps
and a \texttt{FINAL ANSWER} line; the step list follows the description of EverMemOS's
chain-of-thought prompt \cite{evermemos} in the LoCoMo audit \cite{audit}, written in our own words.
In every answer prompt \texttt{\{episodic\}} receives Becos's retrieved context or the whole
conversation, and \texttt{\{semantic\}} is left empty.
\paragraph{Short answer prompt (Nemori).}
\begin{Verbatim}[breaklines,fontsize=\footnotesize]
You are an intelligent memory assistant tasked with retrieving accurate information from conversation memories.

# CONTEXT:
You have access to memories from two speakers in a conversation. These memories contain
timestamped information that may be relevant to answering the question.

# INSTRUCTIONS:
1. Carefully analyze all provided memories from both speakers
2. Pay special attention to the timestamps to determine the answer
3. If the question asks about a specific event or fact, look for direct evidence in the memories
4. If the memories contain contradictory information, prioritize the most recent memory
5. If there is a question about time references (like "last year", "two months ago", etc.),
   calculate the actual date based on the memory timestamp. For example, if a memory from
   4 May 2022 mentions "went to India last year," then the trip occurred in 2021.
6. Always convert relative time references to specific dates, months, or years. For example,
   convert "last year" to "2022" or "two months ago" to "March 2023" based on the memory
   timestamp. Ignore the reference while answering the question.
7. Focus only on the content of the memories from both speakers. Do not confuse character
   names mentioned in memories with the actual users who created those memories.
8. The answer should be less than 5-6 words.

# APPROACH (Think step by step):
1. First, examine all memories that contain information related to the question
2. Examine the timestamps and content of these memories carefully
3. Look for explicit mentions of dates, times, locations, or events that answer the question
4. If the answer requires calculation (e.g., converting relative time references), show your work
5. Formulate a precise, concise answer based solely on the evidence in the memories
6. Double-check that your answer directly addresses the question asked
7. Ensure your final answer is specific and avoids vague time references

Episodic Memories:
{episodic}

Semantic Memories:
{semantic}

Question: {question}

Answer:
\end{Verbatim}
\paragraph{7-step answer prompt (ours); identical to the short prompt except instruction 8.}
\begin{Verbatim}[breaklines,fontsize=\footnotesize]
You are an intelligent memory assistant tasked with retrieving accurate information from conversation memories.

# CONTEXT:
You have access to memories from two speakers in a conversation. These memories contain
timestamped information that may be relevant to answering the question.

# INSTRUCTIONS:
1. Carefully analyze all provided memories from both speakers
2. Pay special attention to the timestamps to determine the answer
3. If the question asks about a specific event or fact, look for direct evidence in the memories
4. If the memories contain contradictory information, prioritize the most recent memory
5. If there is a question about time references (like "last year", "two months ago", etc.),
   calculate the actual date based on the memory timestamp. For example, if a memory from
   4 May 2022 mentions "went to India last year," then the trip occurred in 2021.
6. Always convert relative time references to specific dates, months, or years. For example,
   convert "last year" to "2022" or "two months ago" to "March 2023" based on the memory
   timestamp. Ignore the reference while answering the question.
7. Focus only on the content of the memories from both speakers. Do not confuse character
   names mentioned in memories with the actual users who created those memories.
8. Before answering, reason through these steps and write them out:
   Step 1 - Relevant memories: list every memory that bears on the question, with its timestamp.
   Step 2 - Key details: pull out the names, places, numbers, dates and frequencies they give.
   Step 3 - Connections: link the same people, things and events across memories; note what
            follows from them together.
   Step 4 - Time: turn every relative time reference into a date, month or year using the
            memory's own timestamp.
   Step 5 - Conflicts: if memories disagree, say which one wins and why (most recent first).
   Step 6 - Checklist: confirm the answer keeps every specific name, number and detail asked for.
   Step 7 - Answer: say how you will phrase the answer.
   Then write "FINAL ANSWER:" followed by a concise answer that keeps all specific details.

# APPROACH (Think step by step):
1. First, examine all memories that contain information related to the question
2. Examine the timestamps and content of these memories carefully
3. Look for explicit mentions of dates, times, locations, or events that answer the question
4. If the answer requires calculation (e.g., converting relative time references), show your work
5. Formulate a precise, concise answer based solely on the evidence in the memories
6. Double-check that your answer directly addresses the question asked
7. Ensure your final answer is specific and avoids vague time references

Episodic Memories:
{episodic}

Semantic Memories:
{semantic}

Question: {question}

Answer:
\end{Verbatim}
\paragraph{Judge system message (Nemori).}
\begin{Verbatim}[breaklines,fontsize=\footnotesize]
You are an expert grader that determines if answers to questions match a gold standard answer.
\end{Verbatim}
\paragraph{Judge prompt (Nemori).}
\begin{Verbatim}[breaklines,fontsize=\footnotesize]
Your task is to label an answer to a question as ’CORRECT’ or ’WRONG’. You will be given the following data:
    (1) a question (posed by one user to another user),
    (2) a ’gold’ (ground truth) answer,
    (3) a generated answer
which you will score as CORRECT/WRONG.

The point of the question is to ask about something one user should know about the other user based on their prior conversations.
The gold answer will usually be a concise and short answer that includes the referenced topic, for example:
Question: Do you remember what I got the last time I went to Hawaii?
Gold answer: A shell necklace
The generated answer might be much longer, but you should be generous with your grading - as long as it touches on the same topic as the gold answer, it should be counted as CORRECT.

For time related questions, the gold answer will be a specific date, month, year, etc. The generated answer might be much longer or use relative time references (like "last Tuesday" or "next month"), but you should be generous with your grading - as long as it refers to the same date or time period as the gold answer, it should be counted as CORRECT. Even if the format differs (e.g., "May 7th" vs "7 May"), consider it CORRECT if it's the same date.

Now it’s time for the real question:
Question: {question}
Gold answer: {gold_answer}
Generated answer: {generated_answer}

First, provide a short (one sentence) explanation of your reasoning, then finish with CORRECT or WRONG.
Do NOT include both CORRECT and WRONG in your response, or it will break the evaluation script.

Just return the label CORRECT or WRONG in a json format with the key as "label".
\end{Verbatim}


\begin{thebibliography}{99}
\bibitem{locomo} A.~Maharana et al. Evaluating Very Long-Term Conversational Memory of LLM Agents.
  arXiv:2402.17753, 2024.
\bibitem{mem0} P.~Chhikara et al. Mem0: Building Production-Ready AI Agents with Scalable Long-Term
  Memory. arXiv:2504.19413, 2025.
\bibitem{nemori} W.~Ma, J.~Nan, W.~Wu, Y.~Chen. What Deserves Memory: Adaptive Memory Distillation
  for LLM Agents (Nemori). arXiv:2508.03341v4, 2026.
\bibitem{nemoricode} nemori-ai/nemori, commit d2a6dff (MIT). \url{https://github.com/nemori-ai/nemori}
\bibitem{msft} A.~Gupta, D.~Han, D.~Madrigal Diaz, W.-N.~Chen, M.~Das, V.~R\"uhle, S.~Rajmohan. Long-Term Agent Memory through Complementary Curation and
  Consolidation. arXiv:2610.39790, 2026.
\bibitem{audit} dial481. locomo-audit: Full audit of the LoCoMo benchmark.
  \url{https://github.com/dial481/locomo-audit}
\bibitem{evermemos} EverMind-AI. EverMemOS evaluation prompts. \url{https://github.com/EverMind-AI/EverMemOS}
\bibitem{memgpt} C.~Packer, S.~Wooders, K.~Lin, V.~Fang, S.~G.~Patil, I.~Stoica, J.~E.~Gonzalez.
  MemGPT: Towards LLMs as Operating Systems. arXiv:2310.08560, 2023.
\bibitem{zep} P.~Rasmussen, P.~Paliychuk, T.~Beauvais, J.~Ryan, D.~Chalef. Zep: A Temporal Knowledge
  Graph Architecture for Agent Memory. arXiv:2501.13956, 2025.
\bibitem{amem} W.~Xu, Z.~Liang, K.~Mei, H.~Gao, J.~Tan, Y.~Zhang. A-MEM: Agentic Memory for LLM Agents.
  NeurIPS 2025. arXiv:2502.12110.
\bibitem{msc} J.~Xu, A.~Szlam, J.~Weston. Beyond Goldfish Memory: Long-Term Open-Domain Conversation.
  ACL 2022, pp.~5180--5197.
\bibitem{longmemeval} D.~Wu, H.~Wang, W.~Yu, Y.~Zhang, K.-W.~Chang, D.~Yu. LongMemEval: Benchmarking
  Chat Assistants on Long-Term Interactive Memory. ICLR 2025. arXiv:2410.10813.
\bibitem{lostmiddle} N.~F.~Liu, K.~Lin, J.~Hewitt, A.~Paranjape, M.~Bevilacqua, F.~Petroni, P.~Liang.
  Lost in the Middle: How Language Models Use Long Contexts. TACL 12:157--173, 2024.
\bibitem{ruler} C.-P.~Hsieh, S.~Sun, S.~Kriman, S.~Acharya, D.~Rekesh, F.~Jia, Y.~Zhang, B.~Ginsburg.
  RULER: What's the Real Context Size of Your Long-Context Language Models? COLM 2024.
  arXiv:2404.06654.
\bibitem{judge} L.~Zheng, W.-L.~Chiang, Y.~Sheng, S.~Zhuang, Z.~Wu, Y.~Zhuang, Z.~Lin, Z.~Li, D.~Li,
  E.~P.~Xing, H.~Zhang, J.~E.~Gonzalez, I.~Stoica. Judging LLM-as-a-Judge with MT-Bench and Chatbot
  Arena. NeurIPS 2023 Datasets and Benchmarks. arXiv:2306.05685.
\end{thebibliography}
\end{document}